\documentclass[letterpaper]{article} 
\usepackage{aaai2026}  
\usepackage{times}  
\usepackage{helvet}  
\usepackage{courier}  
\usepackage[hyphens]{url}  
\usepackage{graphicx} 
\usepackage{natbib}  
\usepackage{caption} 
\usepackage{algorithm}
\usepackage{algorithmic}

\usepackage{amsmath,amssymb}

\usepackage{multirow}
\usepackage{tikz}
\usepackage{pgfplots}
\usepackage{pgfkeys}
\usetikzlibrary{shadings}
\usepackage{subcaption}
\usepackage{adjustbox}
\pgfplotsset{compat=1.18}
\usetikzlibrary{calc}
\usetikzlibrary{positioning}

\usepackage{amsthm}
\usepackage{cleveref}
\newtheorem{definition}{Definition}
\newtheorem{theorem}{Theorem}
\newtheorem{cl}{Claim}

\newtheorem{corollary}{Corollary}

\usepackage{newfloat}
\usepackage{listings}
\DeclareCaptionStyle{ruled}{labelfont=normalfont,labelsep=colon,strut=off} 
\floatstyle{ruled}
\newfloat{listing}{tb}{lst}{}
\floatname{listing}{Listing}
\title{FNBT: Full Negation Belief Transformation for Open-World Information Fusion Based on Dempster-Shafer Theory of Evidence}

\author{
    Meishen He\textsuperscript{\rm 1,*},
    Wenjun Ma\textsuperscript{\rm 2,3,*,\textdagger},
    Huijun Yue\textsuperscript{\rm 4},
    Jiao Wang\textsuperscript{\rm 5},
    Xiaoma Fan\textsuperscript{\rm 6}
}
\affiliations{
    \textsuperscript{\rm 1}School of Artificial Intelligence, South China Normal University\\
    \textsuperscript{\rm 2}Aberdeen Institute of Data Science and Artificial Intelligence, South China Normal University\\
    \textsuperscript{\rm 3}School of Computer Science, South China Normal University\\
    \textsuperscript{\rm 4}Otolaryngology, Sun Yat-sen University\\
    \textsuperscript{\rm 5}School of public health, Sun Yat-sen University\\
    \textsuperscript{\rm 6}College of Big Data and Internet, Shenzhen Technology University\\
    * These authors contributed equally to this work.\\
    \textdagger\ Corresponding author: mawenjun@scnu.edu.cn
}

\usepackage{bibentry}

\begin{document}

\maketitle

\begin{abstract}
The Dempster-Shafer theory of evidence has been widely applied in the field of information fusion under uncertainty. Most existing research focuses on combining evidence within the same frame of discernment. However, in real-world scenarios, trained algorithms or data often originate from different regions or organizations, where data silos are prevalent. As a result, using different data sources or models to generate basic probability assignments may lead to heterogeneous frames, for which traditional fusion methods often yield unsatisfactory results. To address this challenge, this study proposes an open-world information fusion method, termed Full Negation Belief Transformation (FNBT), based on the Dempster-Shafer theory. More specially, a criterion is introduced to determine whether a given fusion task belongs to the open-world setting. Then, by extending the frames, the method can accommodate elements from heterogeneous frames. Finally, a full negation mechanism is employed to transform the mass functions, so that existing combination rules can be applied to the transformed mass functions for such information fusion. Theoretically, the proposed method satisfies three desirable properties, which are formally proven: mass function invariance, heritability, and essential conflict elimination. Empirically, FNBT demonstrates superior performance in pattern classification tasks on real-world datasets and successfully resolves Zadeh's counterexample, thereby validating its practical effectiveness.
\end{abstract}



\section{Introduction}

With the rapid development of big data and artificial intelligence, multi-source information fusion has emerged as a crucial technique for enhancing system robustness and improving decision-making reliability \citep{MSIFT2025,DBLP:journals/inffus/ZhangZL25,OUNOUGHI2023}. Among the various tools, the Dempster-Shafer theory of evidence, proposed by \citet{Dempster1967} and further developed by \citet{shafer1976}, has gained prominence for its capacity to handle uncertainty and imprecision, as well as its ability to integrate evidence from multiple, possibly conflicting sources \citep{IEEE2025,FEI2025,LI2024}. Its applications span a wide array of domains, including decision making \citep{DBLP:journals/asc/WangYYL24,GEJS/Xiao/TSMCS2023,DBLP:journals/ijis/MaLLMJM19}, recommendation systems \citep{DBLP:journals/mta/WangQ24}, fault diagnosis \citep{FaultDiagnosis2023,DBLP:journals/inffus/LiuLXS23,DBLP:journals/eswa/PanGDC22}, and target recognition \citep{DBLP:journals/air/Liu23}.

\begin{figure}
    \centering
    \includegraphics[width=0.95\linewidth]{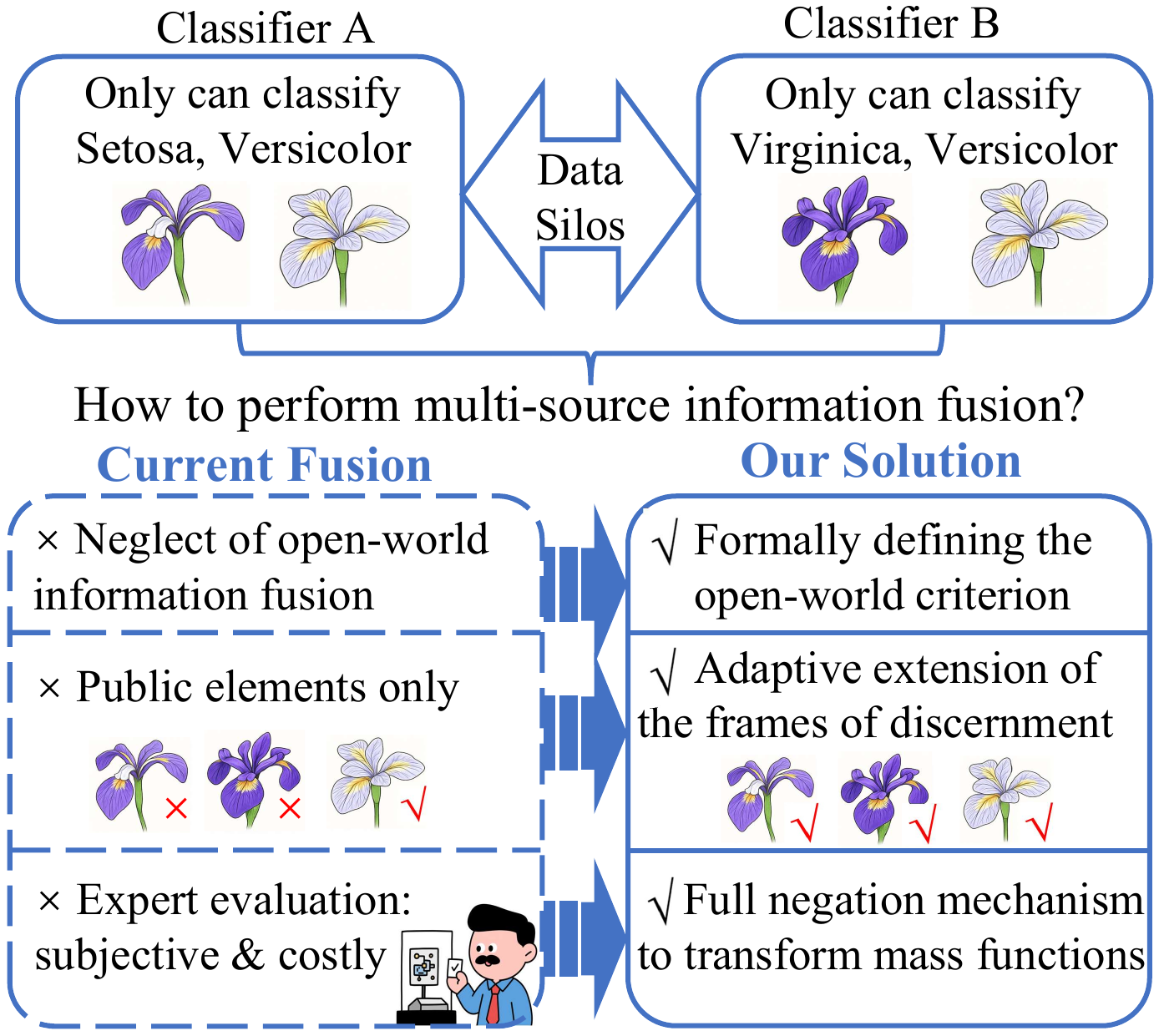}
    \caption{Example of open-world information fusion.}
    \label{fig:Motivation}
\end{figure}

Despite the notable success of Dempster-Shafer theory, there still exist some fundamental limitations for information fusion, especially when the evidence originates from different sources with heterogeneous frames. 
In real-world applications, for the reason of trade secrets, data silos, etc., different subjects may employ different data to train the algorithms, which leads to the fact that the types of samples that can be recognized by different algorithms are often different even for the same task. 
As shown in Figure~\ref{fig:Motivation}, a user has access to two classifiers. Classifier A was trained on data that contains only Setosa and Versicolor. Classifier B was trained on a different dataset from another region that contains only Versicolor and Virginica. Since classifiers A and B were developed by different companies, data silos prevent retraining or data sharing. In this setting, the user aims to determine the class of a given sample among all three possible types (Setosa, Versicolor, Virginica) and must therefore rely on information fusion methods to reach a reliable decision. According to Dempster-Shafer theory of evidence, the corresponding frames for classifiers A and B are $\Theta_A$ = \{Setosa, Versicolor\}, $\Theta_B$ = \{Versicolor, Virginica\}. These frames are clearly heterogeneous, which means that neither frame can encompass the identification of all possibilities for outcomes. As a result, adopting either frame as the frame of discernment would lead to recognition results that are, in effect, out-of-distribution. Therefore, we call the information fusion problem over heterogeneous frames, which is common in practice, an open-world information fusion problem.

Obviously, existing methods fail to effectively address the open-world information fusion problem for three main reasons. First, such problems are often overlooked by existing research, and clear criteria to distinguish open-world from traditional information fusion are lacking. Second, most existing approaches \citep{Xiao/TKDE2025,TKDE2023Xiao,shafer1976} either restrict fusion to the intersecting elements of heterogeneous frames or select one of the frames as the frame of discernment—both of which can lead to significant information loss. Third, when evidence is highly conflicting, belief reassignment or discounting usually hinges on expert evaluation, which is inherently subjective and impractical for end users.

To overcome these limitations, we propose a novel method named Full Negation Belief Transformation (FNBT), designed explicitly for open-world information fusion. First, a criterion is proposed to determine whether a given fusion task is an open-world information fusion problem based on essential conflict. Then, FNBT addresses frame heterogeneity by extending the frame to include elements related to focal elements from heterogeneous frames. Further, we apply a full negation transformation to map mass functions onto the extended effective frame, allowing traditional D-S fusion rules to operate directly in the new frame. FNBT is an objective and adaptive method that requires no expert tuning. It satisfies three critical properties: mass function invariance, heritability, and essential conflict elimination. Experimental results on real-world classification tasks confirm the effectiveness of our method, demonstrating superior performance compared to existing methods.

In summary, this work advances the state-of-the-art in information fusion by introducing a theoretically grounded and practically effective solution for open-world scenarios with heterogeneous frames. We highlight the key contributions below.

\begin{itemize}
    \item \textbf{A practical criterion for distinguishing the open-world information fusion problem:} Aiming at the information fusion problem over heterogeneous frames, this study proposes an open-world criterion grounded in essential conflict, and points out that most existing information fusion methods cannot address the open-world information fusion problem through theoretical analysis.
    \item \textbf{An effective belief transformation method for open-world information fusion:} Based on the extended frame and the full negation mechanism, our FNBT method effectively solves the open-world information fusion problem with desirable properties, including mass function invariance, heritability, and essential conflict elimination.
    \item \textbf{Superior performance across diverse evaluations:} Experiments on pattern classification demonstrate our method’s clear advantage over existing methods in open-world information fusion.
\end{itemize}

\section{Preliminaries}

\begin{definition} \citep{shafer1976} \label{def:mass}
Let $\Theta$ be a set of mutually exclusive and exhaustive elements, called a frame of discernment (or simply a frame). Function $m: 2^{\Theta} \rightarrow [0,1]$ is a mass function, also called a basic probability assignment if: 
\begin{equation}
   m(\emptyset) = 0~and~\sum_{A \subseteq \Theta} m(A) = 1. 
\end{equation}
A subset $A \subseteq\Theta$ satisfying the condition of $m(A)>0$ is called a \textit{focal element}. With respect to m, belief function (Bel) and plausibility function (Pl) are defined as follows:
\begin{align}
&Bel(A) = \sum_{B\subseteq A}m (B), \label{eq:bel}\\
&Pl(A) = \sum_{B\cap A\neq\phi} m (B). \label{eq:pl}
\end{align}
\end{definition}

$Bel(A)$ quantifies the total support given to $A$, while $Pl(A)$ reflects the extent to which $A$ is not contradicted.

\begin{definition} \citep{shafer1976} \label{def:combination-rule}
The combined mass function of $m_1$ and $m_2$ obtained following Dempster’s combination rule is defined as follows:
\begin{equation}
m_{1,2}(x) = 
\begin{cases}
0 & \text{if } x = \emptyset, \\
\dfrac{\displaystyle\sum_{A \cap B = x} m_1(A)m_2(B)}{1 - k_{1,2}} & \text{if $x \neq \emptyset$}.
\end{cases}
\end{equation}
with normalization constant
\begin{equation}
	k_{1,2}=\sum\limits_{A\bigcap B= \emptyset}m_1(A)m_2(B).
\end{equation}
\end{definition}
Here, $k_{1,2}$ is called a classical conflict coefficient, measuring the degree of conflict between pieces of evidence. Specifically, $k_{1,2} = 0$ means that $m_1$ is consistent with $m_2$, while $k_{1,2} = 1 $ implies that there is a total conflict between them. Consequently, the Dempster's combination rule can be applied only if $k < 1$.

\section{Related Work}

Currently, information fusion methods based on evidence theory can be broadly classified into three main categories.

The first type of method proposes combination rules aimed at determining which conflicting state the mass value should be assigned to and how this assignment should be carried out.
The classic Dempster’s combination rule \citep{shafer1976} belongs to this category. However, applying it directly to the open-world setting leads to counterintuitive fusion results, as will be shown in Theorems~\ref{theorem_PlausibilityAbsolutization}--\ref{thm:assertion-closure}.
\citet{DBLP:journals/isci/Yager87a} argues that the mass value of a partial conflict represents ignorance and suggests assigning the mass value of the conflict to the universal set. However, a larger mass value assigned to the universal set implies a higher degree of ignorance.
\citet{DBLP:journals/ci/DuboisP88b} assigns mass value to the union of conflicting focal elements, thereby avoiding direct discarding of conflict information.
However, this rule is neither associative nor quasi-associative, and cannot explicitly describe elements outside the frame.

The goal of the second category of methods is to reconstruct the original mass function based on different assumptions.
The weighted averaging methods aim to weigh the mean of the original mass functions. \citet{DBLP:journals/dss/Murphy00} assigns equal weights to each piece of evidence and then uses Dempster's combination rule for merging. Accordingly, various methods with different types of weight measurement have been proposed, such as \citet{deng2015}'s entropy. In Xiao's research \citep{GEJS/Xiao/TSMCS2023,TKDE2023Xiao,Xiao/TKDE2025,DBLP:journals/apin/ZhuX23}, the divergence or distance metric is often used to express differences between evidence.
The discounting methods \citep{DBLP:journals/isci/ZhaoJS16,shafer1976,DBLP:journals/ci/DuboisP88b,smets2000} consider the reliability of evidence to reconstruct the original mass function, such as the classical method \citep{shafer1976,DBLP:journals/ci/DuboisP88b}, convex combination of sources \citep{smets2000}, inconsistent measurements \citep{DBLP:journals/isci/ZhaoJS16}, contextual discounting \citep{INFORMATIONFUSION2025HUANG}, static discounting coefficient \citep{SOFTCOMPUTING2023QIANG} and the composite discount factors \citep{DBLP:journals/inffus/LiuLXS23}. Applying discounting methods to resolve open-world conflicts often requires experts to manually discount two heterogeneous models, which is subjective and costly. Furthermore, these methods, which are restricted to operating within a single frame, can only select one of the frames in the open-world setting, thereby ignoring certain possibilities.

The third category of methods is based on the open world.
\citet{DBLP:journals/ai/SmetsK94} assign the conflict mass to the empty set in Transferable Belief Model (TBM), to represent the unknown external category. As the number of samples increases, the mass assigned to the empty set tends to increase, but the user is still not informed about which new categories $\emptyset$ specifically points to, and the explainability is insufficient.
In Generalized Evidence Theory \citep{deng2015}, $\emptyset$ can also be a focal element or represent the union of focal elements that are outside the given frame. 
\citet{mGCR} argue that the mass assigned to the empty set should not be assigned to conflict coefficient K, and therefore propose the modified generalized combination rule (mGCR). 
These open-world methods concentrate all conflict on the empty set within a single frame, leaving them unable to fuse evidence from heterogeneous frames or identify which external classes actually emerge.

Existing approaches are inherently unable to tackle open-world information fusion, yet this scenario is commonplace in practice—making an urgently needed solution.

\section{Methodology}

In this section, we propose the FNBT method for solving the open-world information fusion problem. Specifically, we propose a criterion to determine whether a fusion task lies within the open world, and then analyze the adverse effects that arise when Dempster’s rule is applied improperly in this context. Based on this, the proposed FNBT handles the evidence combination challenge over heterogeneous frames by extending the frames and the full negation mechanism.

\subsection{Open-world Criterion}
\label{subsec-opneworld}

Building on the concept of essential conflict~\citep{DBLP:conf/icaart/MaZJ21}, we generalize and extend it to define a formal criterion that identifies when an information fusion task inherently requires an open-world treatment.

\begin{definition}[Open-world Criterion]\label{essential_conflict}
Let $m_1$ and $m_2$ be mass functions defined on frames $\Theta_1$ and $\Theta_2$, respectively. $F_1$ and $F_2$ denote the sets of focal elements of $m_1$ and $m_2$. $\Upsilon_{1,2}\subseteq \Theta_1 \cup \Theta_2$ is a set of essential conflict elements with the mass functions $m_1$ and $m_2$ if and only if for any $\omega \in \Upsilon_{1,2}$, there is $A \in F_i \wedge \omega \in A~(i \in \{1,2\})$ such that:
\begin{equation*}
\forall B \in F_j,~A \cap B = \emptyset \quad (i \neq j~and~i,~j \in\{1,2\}).
\end{equation*}
If $\Upsilon_{1,2}\neq \emptyset$, the evidence combination of $m_1$ and $m_2$ satisfies the open-world criterion; otherwise, if $\Upsilon_{1,2}= \emptyset$, $m_1$ and $m_2$ still in a closed world.
\end{definition}

The implication of the open-world criterion is that when $\Upsilon \neq \emptyset$, there exists at least one focal element whose possible interpretations lie out of distribution with respect to either $\Theta_1$ or $\Theta_2$, rendering independent probability assignments under each frame inadequate. Thus, we characterize such scenarios as open-world cases, requiring a fusion approach beyond conventional Dempster-Shafer methods.

We now show that applying Dempster’s combination rule to such open-world cases introduces two problematic properties.
The first issue concerns the excessive rejection of possible hypotheses due to heterogeneous frames.

\begin{theorem}[Plausibility Absolutization]\label{theorem_PlausibilityAbsolutization}
Assume $m_1$ and $m_2$ are two mass functions over two frames $\Theta_1$ and $\Theta_2$, with a combination result $m_{1,2}$ according to Dempster's combination rule. $Pl_1$, $Pl_2$, and $Pl_{1,2}$ denote the plausibility functions of $m_1$, $m_2$, and $m_{1,2}$, respectively. If the evidence combination of $m_1$ and $m_2$ satisfies the open-world criterion, then there exists $\omega \in \Theta = \Theta_1 \cup \Theta_2$, such that $Pl_1(\omega)>0$ or $Pl_2(\omega)>0$ but $Pl_{1,2}(\omega)=0$.
\end{theorem}

This theorem states that even if $\omega$ is considered possible under at least one mass function, the result of the combination may render it completely implausible. That is, a state judged to be plausible by one source is excluded from consideration in the combination result due to essential conflict, particularly if $\omega \in \Upsilon_{1,2}$. The fused mass function will assign zero mass to all focal elements containing $\omega$.

\begin{theorem}[Assertion of Closure]\label{thm:assertion-closure}
Let the evidence combination of $m_1$ and $m_2$ satisfies the open-world criterion with $\Upsilon_{1,2} \neq \emptyset$, and $m_{1,2}$ be their combination via Dempster's combination rule. Define the consensus set as: $U_{\text{cons}} \triangleq (U_1 \cup U_2) \setminus \Upsilon_{1,2}$ where $U_i$ is the universe of $m_i$'s focal elements. Then: $Pl_{1,2}(U_{\text{cons}}) = 1$ and this remains invariant under further combination with any new evidence.
\end{theorem}

This theorem reveals a fundamental limitation of Dempster's combination rule under open-world assumptions: once the open-world criterion is satisfied, the entire probability measure in the combination result is forcibly assigned to the non-conflicting region $U_{\text{cons}}$, and this outcome is irreversibly fixed in subsequent combinations. This highlights the necessity of designing new fusion mechanisms that can properly handle heterogeneous frames.
See Appendix D for the proofs of Theorems \ref{theorem_PlausibilityAbsolutization}--\ref{thm:assertion-closure}.

\begin{cl}\label{claim_BeliefAbsolutization}
If the evidence combination of two mass functions satisfies the open-world criterion, then the combination using Dempster’s combination rule can render a possible hypothesis either necessarily true or entirely impossible.
\end{cl}


\begin{cl}\label{cl_Uncorrectability}
If the evidence combination of two mass functions satisfies the open-world criterion, then the implausibility of certain states in their combination result cannot be corrected by further combining with any other evidence.
\end{cl}

Analyzing Claims \ref{claim_BeliefAbsolutization} and \ref{cl_Uncorrectability} together, it can be found that if the evidence combination of two mass functions satisfies the open-world criterion, when Dempster's combination rule is applied to combine them, at least one possible state will become impossible or necessary true and will not be corrected by further combination.


\begin{corollary}[Closed-world Consistency]\label{theorem_ClosedWorldSafety}
Let $m_1$ and $m_2$ be mass functions such that $\Upsilon_{1,2} = \emptyset$. Then the combination result under Dempster’s rule does not exhibit plausibility absolutization or assertion of closure.
\end{corollary}

This validates our motivation: in a closed world ($\Upsilon_{1,2} = \emptyset$), classical rules suffice for the task; in the open world ($\Upsilon_{1,2} \neq \emptyset$), a new method for fusing information across heterogeneous frames is required.
Note that if $\Upsilon_{1,2} = \Theta_1 \cup \Theta_2$, then the two mass functions are in complete conflict and Dempster's combination rule is undefined (i.e., normalization denominator is zero). Such extreme cases are excluded from our discussion, as they represent total contradiction rather than open-world fusion.


\subsection{Full Negation Belief Transformation}

\begin{figure}[t]
    \centering
    \includegraphics[width=0.9\linewidth]{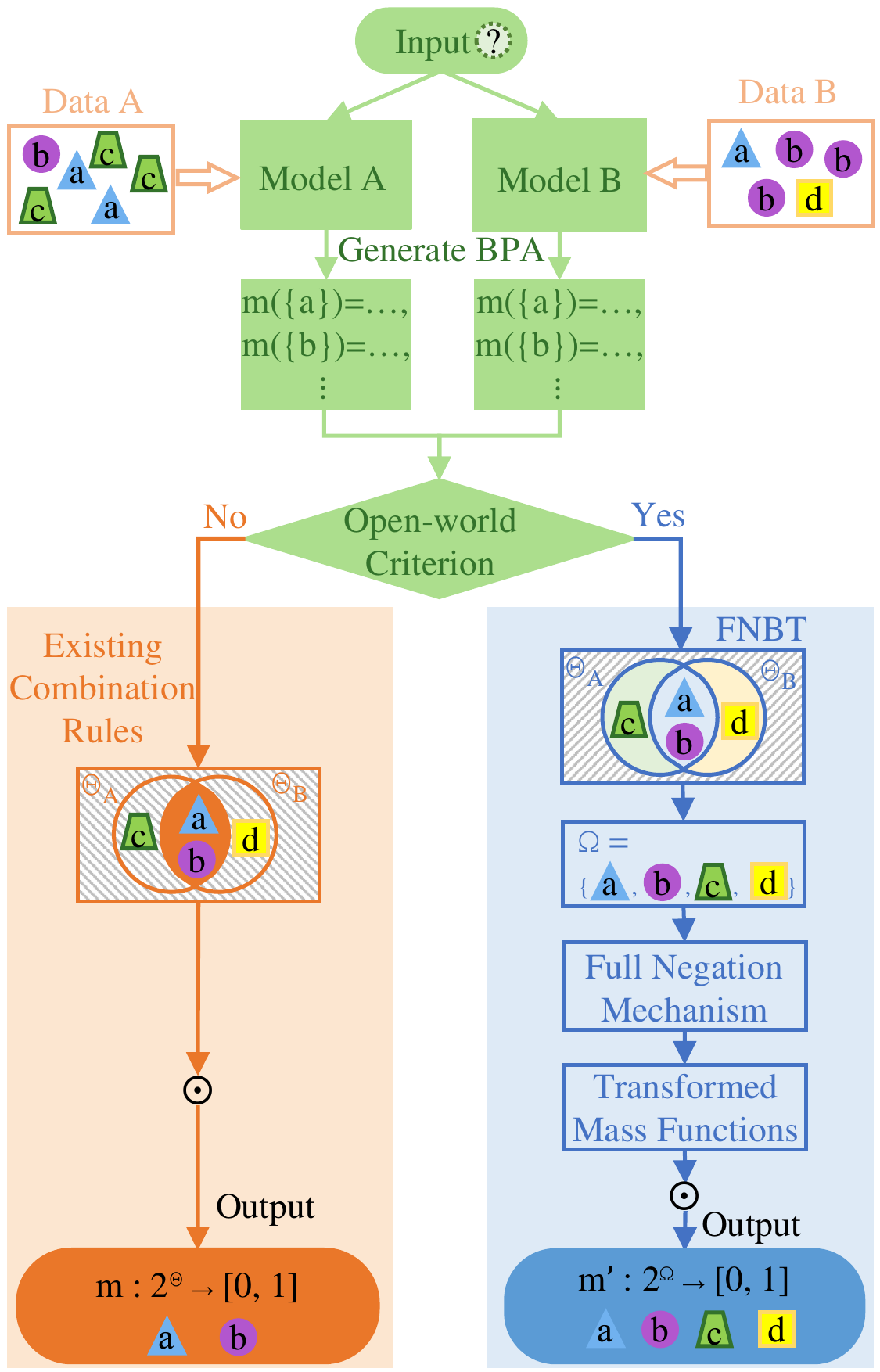}
    \caption{Comparison between the existing combination rules and the proposed FNBT method}
    \label{fig:flowchart}
\end{figure}

The Full Negation Belief Transformation (FNBT) method addresses the challenge of open-world information fusion by extending the frames and leveraging set-theoretic principles to reinterpret basic probability assignments.

Let $m_1$ and $m_2$ be mass functions defined on frames $\Theta_1$ and $\Theta_2$, respectively.

First, according to Definition~\ref{essential_conflict}, if the open-world criterion is not met (i.e., $\Upsilon = \emptyset$), the fusion task falls within a closed-world setting, and Dempster's combination rule can be directly applied. However, if $\Upsilon \neq \emptyset$, indicating an open-world scenario, the following transformation is performed.

Next, to accommodate elements from heterogeneous frames, we extend the frames as defined below.

\begin{definition}
    Assume $m_1$ and $m_2$ are two mass functions over two frames $\Theta_1$ and $\Theta_2$. If the evidence combination of $m_1$ and $m_2$ satisfies the open-world criterion, the effective frame $\Theta_i^*$ for each mass function $m_i$ ($i = 1,2$) is defined as:
    \begin{equation}
    \Theta_i^* = \{ \omega \in \Theta_i \mid \exists A \subseteq \Theta_i,~m_i(A) > 0~\wedge~\omega \in A\}.
    \end{equation}
\end{definition}

In a mass function, any subset assigned zero mass is considered an impossible event. Therefore, the effective frame consists exclusively of the singleton elements that appear in the focal elements—those subsets with nonzero mass. When the effective frames of two mass functions differ, we regard their original frames as heterogeneous.

\begin{definition} \label{def_extended}
    Assume $m_1$ and $m_2$ are two mass functions over two frames $\Theta_1$ and $\Theta_2$. If the evidence combination of $m_1$ and $m_2$ satisfies the open-world criterion with $\Upsilon_{1,2} \neq \emptyset$, the extended effective frame is defined as:
    \begin{equation}
    \Omega = \Theta_1^* \cup \Theta_2^* = \Theta_1^* \cup \Upsilon_{1,2} = \Theta_2^* \cup \Upsilon_{1,2}.
    \end{equation}
\end{definition}

More importantly, the central idea of FNBT lies in the reinterpretation of belief through logical negation. In classical Dempster-Shafer theory, assigning belief to a subset $A \subseteq \Theta$ implicitly expresses negation of its complement, i.e., $\neg A = \Theta \setminus A$. However, in the open-world information fusion problem, the original frame $\Theta$ is incomplete, and such negation must be reinterpreted within the extended effective frame $\Omega$.
For example, let $\Theta_1 = \{\omega_1,~\omega_2,~\omega_3\}$, where $m({\omega_3}) = 0$. The effective frame is $\Theta_1^* = \{\omega_1,~\omega_2\}$. Under this frame, assigning belief to $\{\omega_1\}$ effectively negates $\{\omega_2\}$. If the extended effective frame becomes $\Omega = \{\omega_1,~\omega_2,~\omega_3,~\omega_4\}$, this negation must be revised to support $\{\omega_1,~\omega_3,~\omega_4\}$—namely, all elements except $\omega_2$ in the extended effective frame. This reinterpretation serves as a foundational step in the mass function transformation process employed by FNBT. Express it in axiomatic language:

The transformation applies full negation semantics to each mass function:
\begin{equation}
    m_i(A) = m_i\left( \bigwedge_{\substack{\omega_j \in \Theta_i^* \\ \omega_j \notin A}} \overline{\omega_j} \right),
\quad i = 1, 2,
\end{equation}
where $\overline{\omega_j}$ denotes the negation of element $\omega_j$.

Rewriting this expression in terms of the extended effective frame $\Omega$, we obtain:
\begin{equation}
m_i\left( \bigwedge_{\substack{\omega_j \in \Theta_i^* \\ \omega_j \notin A}} \overline{\omega_j} \right)
= m_i'\left( \Omega \setminus \bigcup_{\substack{\omega_j \in \Theta_i^* \\ \omega_j \notin A}} \{\omega_j\} \right),
\quad i = 1, 2,
\end{equation}
where $m_i'$ denotes the transformed mass function defined over $\Omega$.

After transforming both mass functions, they are combined using Dempster's combination rule.
\begin{equation}
    m_{1,2}' = m_1' \oplus m_2'.
\end{equation}

Dempster's rule is adopted here due to its associative and commutative properties. The FNBT method is also applicable when there are more than two pieces of evidence. Notably, the final fusion step is not limited to Dempster's combination rule and other closed-world combination rules may be used as well.

This process, referred to as \textit{FNBT-Mass}, embeds the logic of full negation into the belief transformation pipeline, thereby ensuring compatibility with open-world settings while mitigating essential conflicts. Furthermore, FNBT supports belief-based and plausibility-based variants, namely \textit{FNBT-Bel} and \textit{FNBT-Pl}, in which belief and plausibility values are computed from the transformed mass functions in accordance with Equations~\eqref{eq:bel} and ~\eqref{eq:pl}.

Figure \ref{fig:flowchart} contrasts the classical combination rules (left) with the proposed FNBT method (right), highlighting the reinterpretation of negation within the extended effective frame $\Omega$, where $\odot$ denotes the existing evidence combination methods. See Appendix G for the pseudocode.

\section{Evalution}

\subsection{Properties Analysis}

\begin{theorem}[Mass Function Invariance]
\label{thm:mass-invariance}
Let $m$ be a mass function defined on the frame $\Theta$, and let $\Upsilon \subseteq \Theta$ be an essential conflict set. After applying the Fully Negation Belief Transformation, the new function $m'$ defined on the extended effective frame $\Omega = \Theta^* \cup \Upsilon$ is a valid mass function, such that:
$m'(\emptyset) = 0$ and $\sum_{A \subseteq \Omega} m'(A) = 1$.
\end{theorem}


FNBT ensures that the resulting transformed function remains a valid mass function under the Dempster-Shafer framework.
This means it preserves key properties such as non-negativity and normalization over the extended effective frame.
As a result, the transformed evidence can still be reliably fused using Dempster's combination rules without breaking the theoretical foundation.


\begin{theorem}[Heritability]
\label{thm:heritability}
When $\Upsilon = \emptyset$, FNBT preserves original mass functions and their combination:
\begin{enumerate}
    \item $\Omega = \Theta_1 = \Theta_2$,
    \item $m_i'(A) = m_i(A)$ for all $A \subseteq \Theta$ and $i = 1,2$,
    \item $m_{1,2}' = m_{1,2}$ (combination results are identical).
\end{enumerate}
\end{theorem}

%
%
%
The theorem demonstrates that the FNBT method is fully compatible with closed-world scenarios and maintains backward compatibility with classical Dempster-Shafer theory. As a result, the step of checking the open-world criterion can be safely omitted in practical applications, simplifying implementation and improving deployment efficiency in industrial settings.

\begin{theorem}[Essential Conflict Elimination]
\label{thm:conflict-elimination}
When $\Upsilon \neq \emptyset$, FNBT eliminates all essential conflicts between $m_1$ and $m_2$:
$\forall A' \in F_1',~\forall B' \in F_2',~A' \cap B' \neq \emptyset$,
where $F_1'$ and $F_2'$ are the focal sets of the transformed mass functions.
\end{theorem}


FNBT eliminates all essential conflicts by expanding each focal element to cover the conflict region.
After transformation, any pair of focal elements from heterogeneous frames always intersect, guaranteeing conflict-free fusion.
This property ensures the effectiveness of the combination process under open-world uncertainty by fundamentally resolving the sources of conflict.

The proofs of Theorems~\ref{thm:mass-invariance}--\ref{thm:conflict-elimination} are given in Appendix E.


\subsection{Experiment}

\subsubsection{Datasets}

\begin{table}
\centering
\begin{tabular}{cccc}
\hline
Datasets & Samples & Classes & Attributes \\ \hline
Iris     & 150     & 3       & 4          \\
Seeds    & 210     & 3       & 7          \\
Wine     & 178     & 3       & 13         \\ \hline
\end{tabular}
\caption{Summarized Information of the Datasets.}
\label{table_datasets}
\end{table}

Real-world datasets from the UCI Machine Learning Repository are utilized, i.e. Iris, Seeds and Wine. Table \ref{table_datasets} provides a summary of each dataset’s characteristics. More details about the datasets can be found in Appendix A.

For each dataset, different attributes are treated as independent sources of information. In this context, the mass generation method from \citet{GEJS/Xiao/TSMCS2023} is employed to generate the initial basic probability assignments.



To emulate heterogeneous frames in an open-world setting, the original training set is split by class into two sample-disjoint subsets whose label sets partially overlap, A and B.  
Let the universal label set be \{class a, class b, class c\}.  
Training set~A contains all samples of class a and half of the samples of class b, corresponding to the frame $\Theta_A = \{\text{class a}, \text{class b}\}$. 
Training set~B comprises the remaining half of class b and all samples of class c, corresponding to $\Theta_B = \{\text{class b}, \text{class c}\}$.  
Consequently, $\Theta_A \cup \Theta_B$ recovers the full label set, while $\Theta_{\mathbf{A}} \cap \Theta_B = \{\text{class b}\}$ supplies the shared class, providing a realistic setting to assess the method's effectiveness in open-world environments.
For the Wine dataset, which exhibits inherent class imbalance, SMOTE is applied during training to mitigate bias.

For each test sample, $n$ pieces of evidence are generated from training set A and another $n$ from training set B ($n$ is the number of features). Within each training set, these $n$ pieces are fused using the classic Dempster’s combination rule to obtain a single piece of evidence $m_A$ and $m_B$, respectively. Note that this intra-subset fusion is performed over the same frame, hence Dempster’s combination rule is appropriate and replaceable if needed. Finally, $m_A$ and $m_B$ are combined with the proposed FNBT method or other comparative methods.

We used accuracy ($\pm$ std) and marco-F\textsubscript{1} ($\pm$ std) to evaluate the effects of the experiments. 

Specific experimental details—code, hardware/software and metrics—are provided in the appendix.

\subsubsection{Model Performance}

In this section, the effectiveness and practicability of the proposed FNBT method are evaluated through a pattern classification task.

\begin{table*}
\centering
\begin{tabular}{ccccccc}
\hline
\multirow{2}{*}{\textbf{Dataset}} & \multicolumn{2}{c}{\textbf{FNBT-Mass}} & \multicolumn{2}{c}{\textbf{FNBT-Bel}} & \multicolumn{2}{c}{\textbf{FNBT-Pl}} \\
& Accuracy & $\mathrm{Marco-F_1}$ & Accuracy & $\mathrm{Marco-F_1}$ & Accuracy & $\mathrm{Marco-F_1}$  \\ \hline
Iris & 90.00 ± 4.47 & 89.70 ± 4.67 & 90.00 ± 4.47 & 89.70 ± 4.67   & \textbf{93.33 ± 4.22}          & \textbf{93.18 ± 4.39}         \\
Seeds & 70.48 ± 6.32 & 69.06 ± 6.65 & 85.71 ± 5.63 & 84.81 ± 6.20  & \textbf{86.67 ± 5.13}          & \textbf{85.91 ± 5.60}         \\
Wine & 72.50 ± 10.10 & 71.69 ± 10.47 & 80.97 ± 7.97 & 80.26 ± 8.95  & \textbf{86.05 ± 7.86}          & \textbf{85.88 ± 8.14}         \\ \hline
\end{tabular}
\caption{Pattern Classification Results.}
\label{classification-result}
\end{table*}

A 10-fold cross-validation scheme is used for evaluation.
To handle evidence from training sets A and B, which are defined over heterogeneous frames, three decision strategies under FNBT are explored: FNBT-Mass, FNBT-Bel, and FNBT-Pl. The Mass strategy prioritizes the hypothesis with the highest mass assigned to a singleton focal element; if the maximum mass lies in a multi-element focal set, a random element from that set is selected as the final decision.


Table \ref{classification-result} reports the classification performance of the three FNBT decision strategies across the three datasets. The results show that FNBT-Pl consistently achieved the best performance: 93.33\% ± 4.22 on Iris, 86.67\% ± 5.13 on Seeds, and 86.05\% ± 7.86 on Wine. In particular, when test instances belonged to classes not shared between training sources, the FNBT-Pl strategy, by aggregating the mass of all hypotheses overlapping with the target class, exhibited superior flexibility and inclusiveness. This highlights the FNBT-Pl's ability to better capture and manage uncertainty in open-world scenarios. In contrast, the FNBT-Mass strategy, which directly selects hypothesis with the maximum mass, was more vulnerable to confusion caused by multi-element sets, resulting in the lowest performance. The FNBT-Bel strategy, which considers only strict subsets of hypotheses, was found to be overly conservative under open world conditions.

\subsubsection{Comparison}

\begin{table*}
\centering
\begin{tabular}{ccccccc}
\hline
\multirow{2}{*}{\textbf{Method}} & \multicolumn{2}{c}{\textbf{Iris}}            & \multicolumn{2}{c}{\textbf{Seeds}}          & \multicolumn{2}{c}{\textbf{Wine}}            \\
                        &  Accuracy          & Marco-F\textsubscript{1}        &  Accuracy          & Marco-F\textsubscript{1}        &  Accuracy          & Marco-F\textsubscript{1}         \\ \hline
Dempster's              & 33.33\%          & 16.67\%          & 33.33\%          & 16.67\%          & 38.89\%          & 18.67\%          \\
ETV-MSIF                  & 66.67\%          & 55.56\%          & 66.67\%          & 53.47\%          & 66.67\%          & 65.16\%          \\
Yager's                 & 33.33\%          & 16.67\%          & 33.33\%          & 16.67\%          & 38.89\%          & 18.67\%          \\
TBM                 & 33.33\%          & 23.81\%          & 21.43\%          & 26.09\%          & 27.78\%               & 27.78\%               \\
mGCR                    & 33.33\%          & 16.67\%          & 33.33\%          & 16.67\%          & 38.89\%          & 18.67\%          \\
\textbf{FNBT-Pl}        & \textbf{93.33\%} & \textbf{93.27\%} & \textbf{83.33\%} & \textbf{81.70\%} & \textbf{86.11\%} & \textbf{85.95\%} \\ \hline
\end{tabular}
\caption{Comparison Results.}
\label{table-comparison}
\end{table*}

To further validate the effectiveness of the proposed FNBT method in the open-world information fusion, we conducted comparative experiments against several existing methods. 
In these experiments, 20\% of the samples from each class were randomly selected as the test set, while the remaining 80\% formed the training set.

As summarized in Table~\ref{table-comparison}, results are given as average accuracy and Marco-F\textsubscript{1}. 
Take the accuracy of each comparison method on the Iris dataset as an example. 
The classic Dempster’s rule \citep{shafer1976}, which forcibly normalizes mass by redistributing all the conflicting mass to intersecting focal elements, yields only 33.33\% accuracy.
Yager’s rule \citep{DBLP:journals/isci/Yager87a}, which redirects the conflict to the universal set, avoids normalization issues but leads to overly uncertain decisions, again yielding 33.33\% accuracy.
TBM \citep{DBLP:journals/ai/SmetsK94}, which assigns conflict to the empty set ($m(\emptyset)$) to allow for open-world modeling, suffers from undefined semantics in the decision phase, resulting in subpar performance (33.33\% accuracy). 
mGCR \citep{mGCR}, while theoretically permitting $m(\emptyset)>0$, fails to capture the relationship between heterogeneous sources, and thus performs no better.
ETV-MSIF \citep{Xiao/TKDE2025}, which preprocesses and iteratively fuses conflicting evidence, improves the result to 66.67\%, yet remains confined by the closed-world assumption.

\begin{figure}
    \centering
    \includegraphics[width=0.96\linewidth]{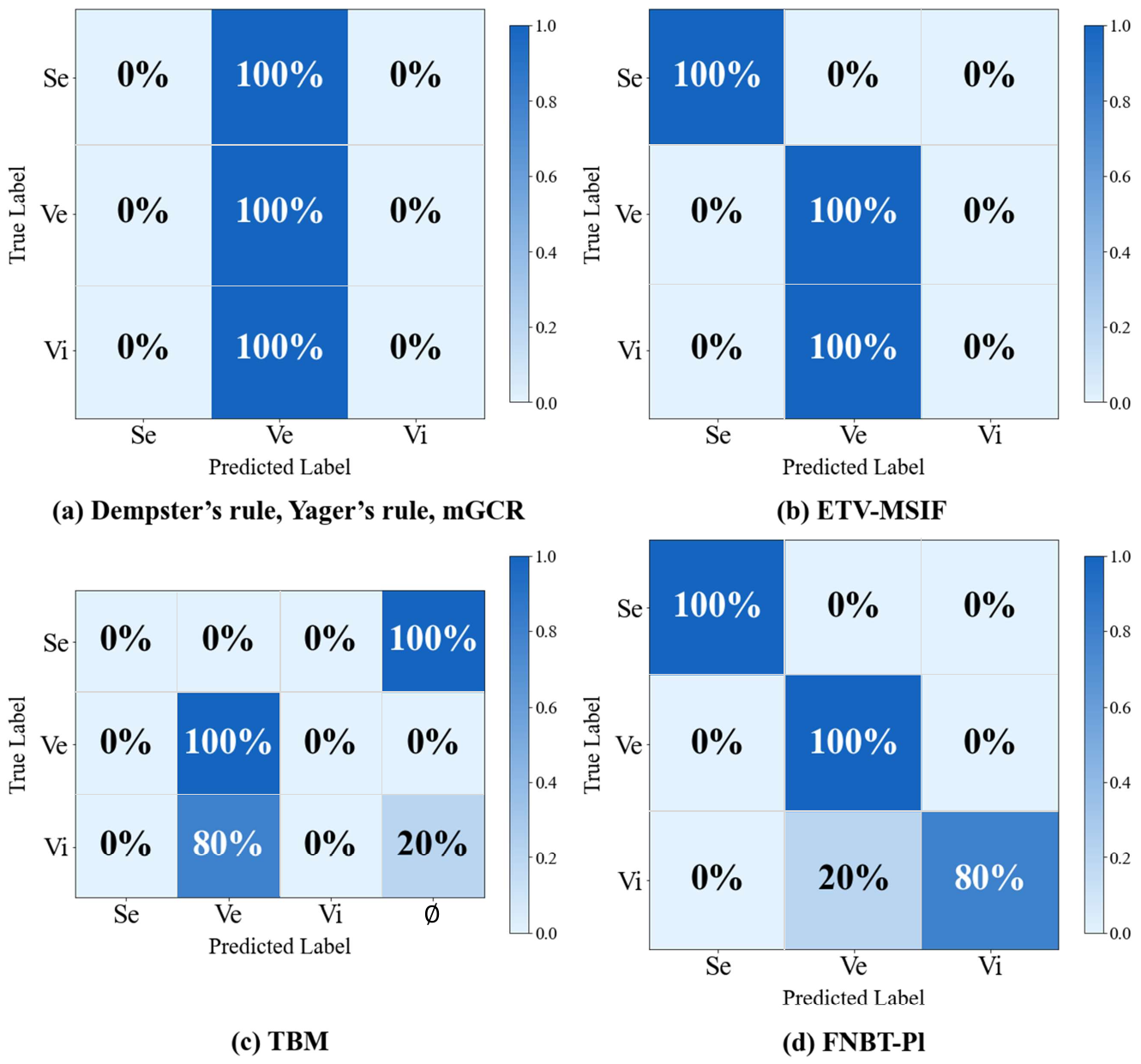}
    \caption{Comparison of Confusion Matrices.}
    \label{fig:conf_matrix_compare}
\end{figure}

%
%

In addition, the experimental performance of these methods \citep{DBLP:journals/cam/ZengX24,DBLP:journals/air/Liu23,DBLP:journals/apin/ZhuX23,DBLP:journals/apin/HuaJ23,SOFTCOMPUTING2023QIANG,DBLP:journals/inffus/LiuLXS23} closely matches that of ETV-MSIF because they fuse the weighted average evidence n-1 times using the Dempster's combination rule, resulting in a final fusion that predominantly favors one of the frames.

In sharp contrast, our FNBT-Pl method significantly outperforms all baselines, achieving 93.33\% accuracy and 93.27\% macro-F\textsubscript{1} on the Iris dataset, with compelling results on Seeds and Wine. These findings underscore the effectiveness of FNBT-Pl in open-world information fusion.

The confusion matrices in Figure~\ref{fig:conf_matrix_compare} provide further insight into class-wise prediction behaviors across different methods.
Dempster’s rule, Yager’s rule, and mGCR (Figure 3a) exhibit extreme bias toward Ve, assigning all samples to this class. As a result, Se and Vi achieve near-zero recall, indicating severe class confusion.
ETV-MSIF (Figure 3b) correctly classifies all Se and Ve samples but completely fails to distinguish Vi. This indicates its limited generalization under heterogeneous frames.
TBM (Figure 3c) introduces an additional decision category for the empty set, yet fails to correctly classify any Se or Vi samples. This underscores the challenge of making semantically meaningful decisions when the semantics of the empty set do not map to any specific class.
In contrast, FNBT-Pl (Figure 3d) achieves near-perfect per-class accuracy: 100\% for Se and Ve, and 80\% for Vi. This clear diagonal structure indicates high precision and recall across all classes, demonstrating FNBT-Pl’s superior discriminative capability.






\subsection{Case Study}
\label{case:zadeh}

We demonstrate the validity of the proposed FNBT method through Zadeh's counterexample \citep{DBLP:journals/aim/Zadeh86}. For completeness, the counterexample is provided in Appendix F. 

\noindent\textbf{Step 1: Open-World Criterion Assessment}\\
Consider mass functions $m_1$ and $m_2$ defined on $\Theta_1 = \Theta_2 = \{a,b,c\}$ with assignments:
\begin{align*}
&m_1(\{a\}) = 0.9,\ m_1(\{b\}) = 0.1,\ m_1(\{c\}) = 0; \\
&m_2(\{a\}) = 0,\ m_2(\{b\}) = 0.1,\ m_2(\{c\}) = 0.9.
\end{align*}

By Definition~\ref{essential_conflict}, the essential conflict set $\Upsilon_{1,2} = \{a,c\}$ is non-empty. It satisfies the open-world criterion, triggering FNBT's open-world information fusion processing.

\noindent\textbf{Step 2: Frame Extension}\\
The effective frames are derived as $\Theta_1^* = \{a,b\}$ (since $m_1(\{c\}) = 0$) and $\Theta_2^* = \{b,c\}$ (since $m_2(\{a\}) = 0$). 
Further, the extended effective frame is constructed as:
\[
\Omega = \Theta_1^* \cup \Theta_2^* = \{a,b,c\}.
\]
All subsequent operations are performed within this frame.

\noindent\textbf{Step 3: Full Negation Mechanism}\\
FNBT reinterprets each mass assignment through logical negation semantics relative to $\Omega$:

\begin{itemize}
    \item \textit{Transformation of $m_1$:} 
    \begin{align*}
        m_1(\{a\}) = 0.9 \mapsto m_1(\{\neg b\}) = 0.9 \mapsto m_1'(\{a,c\}) = 0.9\\
        m_1(\{b\}) = 0.1 \mapsto m_1(\{\neg a\}) = 0.1 \mapsto m_1'(\{b,c\}) = 0.1
    \end{align*}
    
    \item \textit{Transformation of $m_2$:} 
    \begin{align*}
        m_2(\{b\}) &= 0.1 \mapsto m_2(\{\neg c\}) = 0.1 \mapsto m_2'(\{a,b\}) = 0.1\\
        m_2(\{c\}) &= 0.9 \mapsto m_2(\{\neg b\}) = 0.9 \mapsto m_2'(\{a,c\}) = 0.9
    \end{align*}
\end{itemize}


\noindent\textbf{Step 4: Evidence Combination}\\
Applying Dempster's combination rule to combine the transformed mass functions yields the following results:
\begin{align*}
m_{1,2}'(\{a,c\}) &= 0.81,\ m_{1,2}'(\{a\}) = 0.09; \\
m_{1,2}'(\{c\}) &= 0.09,\ m_{1,2}'(\{b\}) = 0.01.
\end{align*}
%

FNBT produces a rationally distributed result: $\{a,c\}$ holds $81\%$ belief, while $\{a\}$ and $\{c\}$ retain $9\%$ support each. Crucially, $\{b\}$ receives only $1\%$ mass -- \textit{not} $100\%$ as in Dempster's rule. This resolves Zadeh's core objection: (i) State $a$ (strongly supported by $m_1$) and $c$ (strongly supported by $m_2$) are preserved in the solution ($\{a,c\}$ dominates). (ii) State $b$, weakly supported by both, is appropriately discounted. (iii) No counterintuitive absolute denial of strongly supported hypotheses.

The case validates FNBT’s efficacy in resolving open-world conflicts via the full-negation mechanism and by extending the heterogeneous frames, aligning with human intuition while maintaining rigorous theoretical foundations.

\section{Conclusion}

This paper addresses the challenging problem of information fusion under heterogeneous frames, a scenario commonly encountered in real-world applications but often overlooked in traditional fusion methods. To bridge this gap, we propose a novel open-world information fusion method named Full Negation Belief Transformation (FNBT). The method first introduces an open-world criterion based on essential conflict to detect when fusion tasks involve heterogeneous frames. Then, it systematically extends the frame to accommodate all relevant elements and applies a full negation mechanism to map the original mass functions to the extended effective frame, thereby enabling the existing combination rules to function effectively in the open-world setting.

FNBT is theoretically sound and satisfies three key properties: mass function invariance, heritability, and essential conflict elimination. Empirical evaluations on real-world classification tasks demonstrate that FNBT significantly outperforms existing fusion methods, highlighting its practical effectiveness and adaptability.

In conclusion, FNBT provides a principled and efficient solution for open-world information fusion on heterogeneous frameworks, pushing the boundaries of D-S theory and offering new possibilities for reliable decision-making in constantly evolving and data-siloed environments.

\bibliography{main}


\end{document}